\documentclass[nopreprintline,12pt,authoryear]{elsarticle}

\usepackage{amssymb}
\usepackage{amsmath}
\usepackage{float}
\usepackage{lineno}
\usepackage{colortbl}  
\usepackage{xcolor}
\usepackage{array}  
\usepackage{pifont}
\usepackage{caption}
\usepackage{tabularx}
\usepackage{orcidlink}
\usepackage{multirow}
\usepackage{url}
\usepackage{booktabs}
\usepackage{threeparttable}
\usepackage{caption}
\newcolumntype{Y}{>{\centering\arraybackslash}X}

\journal{Computers and Electronics in Agriculture}

\begin{document}

\begin{frontmatter}

\title{StomataSeg: Semi-Supervised Instance Segmentation for Sorghum Stomatal Components}


\author[1]{Zhongtian Huang\orcidlink{0009-0005-3399-9695}}
\author[1]{Zhi Chen*\orcidlink{0000-0002-9385-144X}}
\cortext[cor1]{Corresponding author: Zhi Chen}
\ead{uqzhichen@gmail.com}
\author[2]{Zi Huang\orcidlink{0000-0002-9738-4949}}
\author[3]{Xin Yu\orcidlink{0000-0002-0269-5649}}
\author[2]{Daniel Smith\orcidlink{0000-0002-5867-9613}}
\author[2]{Chaitanya Purushothama\orcidlink{0009-0006-9430-8912}}
\author[2]{Erik Van Oosterom\orcidlink{0000-0003-4886-4038}}
\author[2]{Alex Wu\orcidlink{0000-0002-6612-7691}}
\author[4]{William Salter\orcidlink{0000-0003-1653-5192}}
\author[1]{Yan Li\orcidlink{0000-0002-4694-4926}}
\author[2]{Scott Chapman\orcidlink{0000-0003-4732-8452}}



\affiliation[1]{organization={University of Southern Queensland},
            city={Toowoomba},
            postcode={4350}, 
            state={Queensland},
            country={Australia}}
\affiliation[2]{organization={The University of Queensland},
            city={St Lucia},
            postcode={4072}, 
            state={Queensland},
            country={Australia}}
\affiliation[3]{organization={Adelaide University},
            city={Adelaide},
            postcode={5005}, 
            state={SA},
            country={Australia}}            
\affiliation[4]{organization={University of Sydney},
            city={Sydney},
            postcode={2050}, 
            state={NSW},
            country={Australia}}            
            
  
\begin{abstract}
Sorghum is a globally important cereal grown widely in water-limited and stress-prone regions. Its strong drought tolerance makes it a priority crop for climate-resilient agriculture. Improving water-use efficiency in sorghum requires precise characterisation of stomatal traits, as stomata control of gas exchange, transpiration and photosynthesis have a major influence on crop performance. Automated analysis of sorghum stomata is difficult because the stomata are small (often less than 40 µm in length in grasses such as sorghum) and vary in shape across genotypes and leaf surfaces. Automated segmentation contributes to high-throughput stomatal phenotyping, yet current methods still face challenges related to nested small structures and annotation bottlenecks.
In this paper, we propose a semi-supervised instance segmentation framework tailored for analysis of sorghum stomatal components. We collect and annotate a sorghum leaf imagery dataset containing \textbf{11,060} human-annotated patches, covering the three stomatal components (pore, guard cell and complex area) across multiple genotypes and leaf surfaces. To improve the detection of tiny structures, we split high-resolution microscopy images into overlapping small patches. We then apply a pseudo-labelling strategy to unannotated images, producing an additional \textbf{56,428} pseudo-labelled patches. Benchmarking across semantic and instance segmentation models shows substantial performance gains: for semantic models the top mIoU increases from 65.93\% to 70.35\%, whereas for instance models the top AP rises from 28.30\% to 46.10\%.  These results demonstrate that combining patch-based preprocessing with semi-supervised learning significantly improves the segmentation of fine stomatal structures. The proposed framework supports scalable extraction of stomatal traits and facilitates broader adoption of AI-driven phenotyping in crop science.
\end{abstract}

\begin{keyword}
stomata segmentation; multi-class instance segmentation; multi-class semantic segmentation; semi-supervised learning; sorghum leaf stomata dataset; plant phenotyping
\end{keyword}

\end{frontmatter}



\section{Introduction}
\label{sec:introduction}

Sorghum (Sorghum bicolor (L.) Moench) is a globally important crop that plays a key role in agriculture, particularly in arid and semi-arid regions \citep{sanjana2017sorghum, khalifa2023assessment, borrell2021physiology}. It serves as a major cereal and forage resource in areas where water is limited \citep{wondimu2023genome}. Sorghum requires significantly less water than many other cereals and maintains productivity under drought stress and low-input farming conditions \citep{begna2025heterosis, borrell2014drought, borrell2024drought}. These traits make it valuable for food security, livestock feed, and bioenergy supply in vulnerable regions \citep{hossain2022sorghum, mwamahonje2021drought, hammer2019sorghum}. Strengthening analytical tools for sorghum can therefore support breeding programs focused on climate resilience, water-use efficiency, and sustainable agriculture.

Understanding how sorghum responds to challenging environments requires detailed examination of its stomata \citep{khanthavong2022combinational}. Stomata are microscopic pores on plant leaf surfaces that regulate gas exchange, facilitate photosynthesis, and control transpiration. They have a direct impact on plant growth, productivity, and responses to environmental stresses such as drought or elevated atmospheric CO$_2$ levels \citep{lawson2004stomatal, taylor2012photosynthetic}. Given their pivotal role in plant physiology, stomatal traits analysis on stomatal density, size, pore openness, and spatial distribution is critical for understanding plant adaptation strategies, improving crop yield, and breeding varieties with higher environmental resilience \citep{buckley2019stomata, faralli2019exploiting}.

Historically, stomatal analysis has relied heavily on manual microscopy and image-based measurement techniques, which are labour-intensive, subjective, and difficult to scale in high-throughput phenotyping contexts \citep{fetter2019stomatacounter, yang2020crop}. More recently, automated methods leveraging computer vision and deep learning \citep{wei2024plantseg,wei2024benchmarking,wei2025augment,wei2024snap,chen2024cf,chen2021semantics,chen2023zero,chen2025svip,chen2021mitigating,chen2020canzsl,chen2025cluster,chen2025fastedit,chen2025distributed,chen2022gsmflow,chen2020rethinking,zhao2025continual,wang2025discrimination} have been applied to overcome these limitations, which allows more efficient and objective extraction of stomatal traits from large image sets \citep{jayakody2017microscope, casado2020labelstoma, tan2024machine, xie2021optical}.

Several recent studies further illustrate the state of automated stomatal analysis. \citet{gibbs2024application} provided a comprehensive review of deep-learning approaches for stomatal detection and segmentation. They noted the predominance of semantic and bounding-box methods and advocated for more instance-level and multi-class solutions. Advancing the capability for real-time monitoring, \citet{sun2023stomatatracker} introduced StomataTracker, a deep-learning framework designed to reveal circadian rhythms in wheat by processing in-situ video data. More specifically, \citet{thai2024comparative} conducted a comparative analysis of the commonly used models, Mask R-CNN\citep{he2017mask} and YOLOv8\citep{yolov8_ultralytics}. Their study focused on pore segmentation and showed that the two models behave very differently when segmenting the pore area, which is extremely small and often low-contrast in microscopy images. Meanwhile, the StomaVision system developed by \citet{wu2024stomavision} tackles stomatal pore area measurement across several crop species using an instance-segmentation workflow. This work highlights both the promise and persistent challenges of fine-scale segmentation in real-world microscopy images. Collectively, these works underline two major limitations. One is \textbf{the scarcity of multi-class, nested-object segmentation studies for the area of the stomatal complex, and its nested components of guard cell and pore areas}. The other one is \textbf{the high annotation cost required for dense masks}, particularly for classes of small objects such as the pore area.

\begin{figure}[h]
  \centering
  \includegraphics[width=0.95\textwidth]{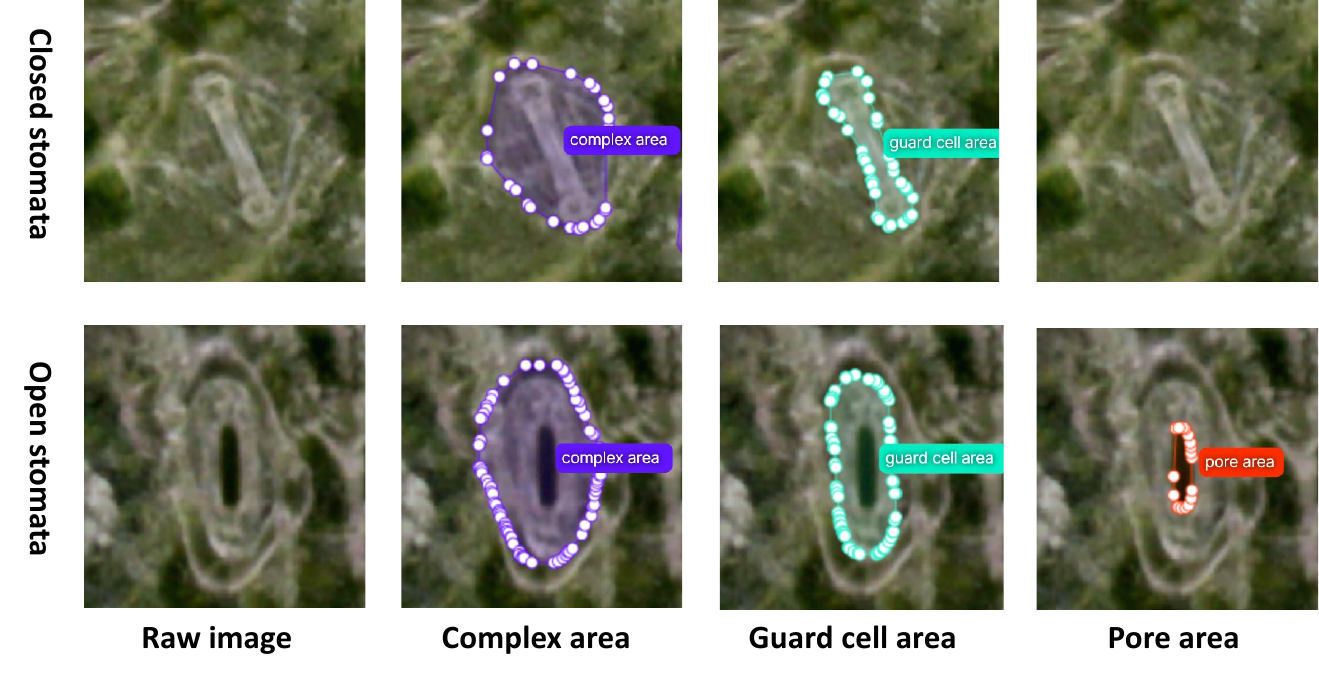}
  \caption{Closed stomata show no visible pore; open stomata exhibit a distinct pore.}
  \label{fig:pore}
\end{figure}

In our segmentation pipeline, each stomatal instance is divided into three biologically meaningful areas as shown in Figure~\ref{fig:pore}, including the stomatal complex area, the guard cell area, and the pore area. The stomatal complex area corresponds to the stomatal complex area previously measured in sorghum and includes both guard cells and the pore \citep{bheemanahalli2021classical}. The guard cell area isolates the paired guard cells, which in grasses such as sorghum regulate opening and closing dynamics under environmental stress \citep{rui2024guard}. The pore area represents the actual opening between guard cells, visible only when stomata are open. Figure~\ref{fig:pore} illustrates closed stomata without a visible pore and open stomata with a clear pore opening. In sorghum, pore dimensions have been linked to conductance responses under changing light and water conditions \citep{battle2024fast}. Sorghum poses particular difficulties for segmentation because natural variation among genotypes and leaf surfaces causes large differences in stomatal density, size, and spacing \citep{hossain2022sorghum}. Moreover, as a grass species, sorghum has dumbbell-shaped guard cells flanked by subsidiary cells, which produce fine structural boundaries and often small or faint pores, which are anatomical traits that complicate fine-grained segmentation of the complex area, guard cell area, and pore area. By applying our segmentation framework to sorghum, we subject it to realistic and demanding phenotyping conditions in a crop of high agricultural relevance.

To overcome the limitations identified in prior work and enable more comprehensive analysis of stomatal architecture, we introduce a unified pipeline for multi-class stomatal instance segmentation in sorghum. This pipeline is designed both to lower the annotation burden and to enhance model performance on small and nested structures. It includes: (a) a patch-based preprocessing strategy that magnifies small stomatal features and mitigates the scaling disadvantages of full-frame microscopy images; (b) a semi-supervised pseudo-labelling process that leverages high-confidence model predictions on unlabelled data to expand the training set without proportional annotation costs; and (c) the development of a multi-class instance segmentation dataset called StomataSeg including 11,060 human-annotated and 56,428 pseudo-labelled image patches. This dataset encompasses the stomatal complex, guard cell, and pore classes across five genotypes and multiple leaf surfaces and regions. With the accompanying dataset provided as an experimental resource, the primary objective is to advance the segmentation capability of the three stomatal classes (i.e. stomatal complex area, guard cell area, and pore area) in order to enable scalable trait extraction and high-throughput phenotyping of stomata, using sorghum (Sorghum bicolor (L.) Moench) as a model crop. 


\begin{figure}[h]
  \centering
  \includegraphics[width=0.9\textwidth]{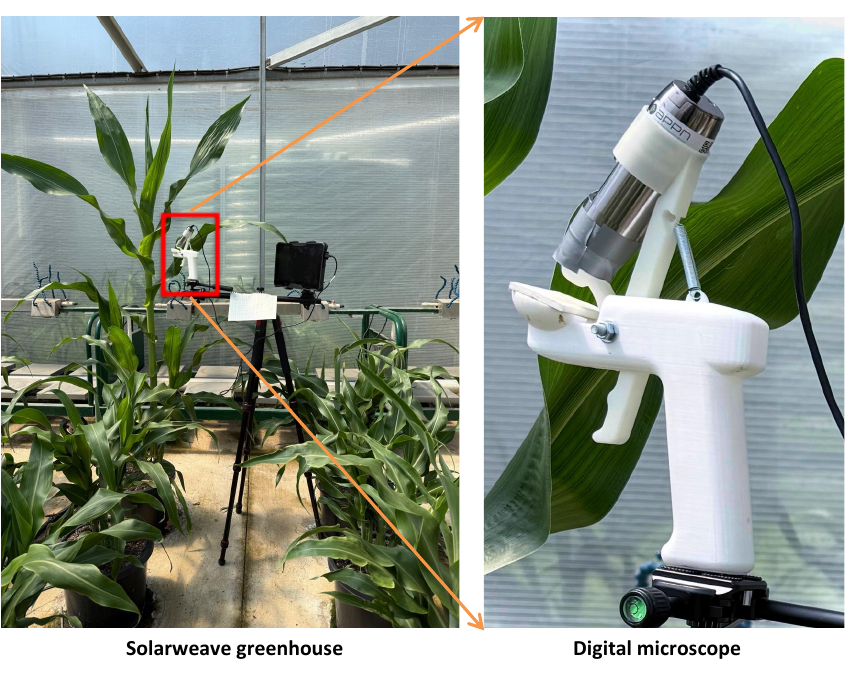}
  \caption{Digital microscopy was performed inside a solarweave greenhouse.}
  \label{fig:digital_microscope}
\end{figure}

\section{Material and methods}
\label{sec:materials_methods}

\subsection{Image acquisition}
\label{subsec:image_acquisition}

All images used for the annotation process were captured from fully expanded leaves of sorghum plants grown in a solarweave greenhouse at the Gatton campus of the University of Queensland, Australia. The images were captured using a high-resolution digital microscope (Dino-Lite Edge/5MP AM7515). We used a 3D printed microscope holder~\citep{chaplin2025smarter} to house the microscope, permitting imaging of leaves from multiple angles. The microscope was attached to a camera tripod and microscope software was run on an accompanying tablet PC. Fully expanded leaves from five genotypes were imaged while still attached to the plants. For each genotype, images were taken from three longitudinal regions of the leaf blade (tip, mid, and base) and from both adaxial (upper) and abaxial (lower) surfaces. This sampling strategy ensures a broad coverage of natural stomatal variation.

As shown in Fig.~\ref{fig:digital_microscope}, microscopy was conducted on intact leaves that had been exposed to ambient greenhouse light conditions prior to microscope attachment. This allowed the stomata to remain in a physiologically active state rather than being dark-adapted, thus ensuring high conductance and enough open stomata. A fixed magnification was used across all samples, producing images at a resolution of 2592 × 1944 pixels in JPEG format. This resolution is sufficient to capture detailed cellular features of stomata. To ensure quality, multiple replicate images were taken from the same leaf portion with only the clearest image being retained. Each image was assigned a structured filename that encoded genotype, replicate, leaf number, surface, and region, providing traceability and facilitating downstream analysis.

We maintained consistency in imaging conditions while preserving the authenticity of stomatal structures on living leaves. This approach results in high-quality microscopy images suitable for manual annotation and pseudo-labels generation.

\subsection{Image sampling and selection}
\label{subsec:sampling}

To capture representative variation in sorghum stomatal traits, images were collected across multiple biological factors. Microscope images were acquired from fully expanded leaves of five genotypes, from both the abaxial (lower) and adaxial (upper) leaf surfaces. Within each surface, this sampling mechanism covered three longitudinal leaf regions (base, middle, tip) to reflect natural gradients in stomatal density and morphology \citep{al2023anatomical}. This design aimed to capture a broad range of natural variation rather than focusing on a single genotype or region.

All images were subjected to quality checks to ensure sharp focus, appropriate illumination, and clear visibility of each stomatal contour. Images that did not meet these criteria were excluded from the subsequent annotation process. Where possible, sampling was balanced across genotype–surface–region combinations, though some variation remained due to biological and practical constraints. Finally, 318 high quality images were assigned for downstream annotation.

\begin{figure}[h]
  \centering
  \includegraphics[width=0.9\textwidth]{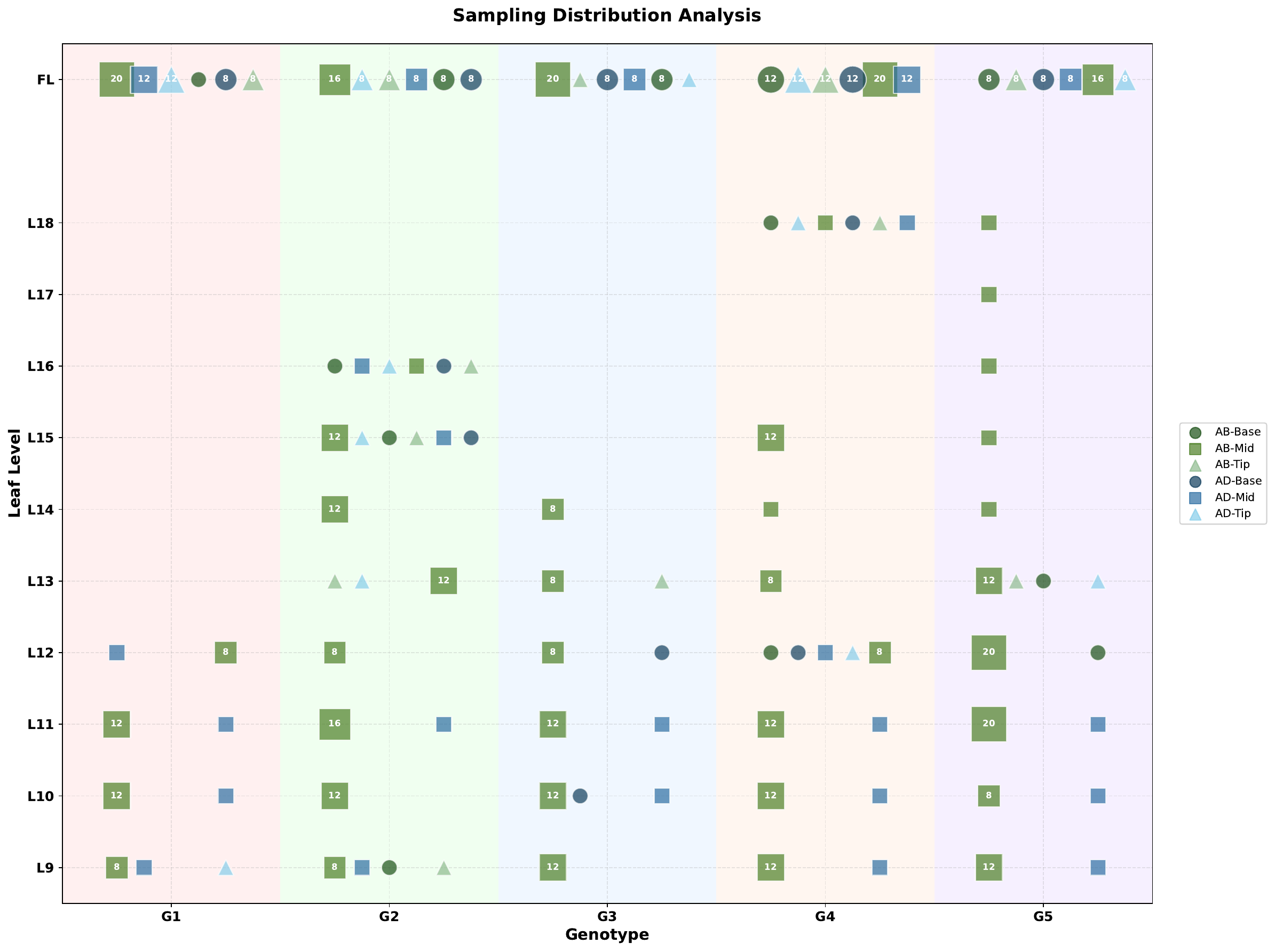}
  \caption{Distribution of the 318 human-annotated original images across genotype, leaf level, surface and region dimensions. Each point denotes a specific genotype–leaf-level combination; colour (green = abaxial, blue = adaxial) indicates leaf surface, shape (circle = base, square = mid, triangle = tip) indicates sampling region, and the number shows image count. Background bands separate the five genotypes. The y-axis tracks developmental progression from L9 (lower leaves) to FL (flag leaf, uppermost fully-emerged leaf).}
  \label{fig:comprehensive_distribution}
\end{figure}

Fig.~\ref{fig:comprehensive_distribution} visualizes the comprehensive distribution of the 318 human-annotated images across all four sampling dimensions: genotype, leaf level, surface, and region. The scatter plot employs a multi-dimensional encoding scheme where each data point represents a specific genotype–leaf level combination. Point colour encodes leaf surface (green for abaxial, blue for adaxial), point shape encoded sampling region (circles for base, squares for mid, triangles for tip), and point size reflects the number of annotated images in that combination. For visual clarity, numerical counts are displayed directly on larger data points, while smaller points indicate lower counts. Background colour bands visually separate the five genotypes (QL12, TX7000, R931945-2-2, SC170-6-8, and SC237-14E) for easy identification.

The y-axis represents developmental progression from L9 (lower leaves, closer to the stem base) through intermediate levels (L10–L18) to FL (flag leaf, the uppermost fully-emerged leaf at the plant apex before reproductive head emergence). This developmental range captures stomatal variation across different growth stages, with concentrated sampling at biologically important levels, particularly L9–L12 for mature tissue and FL for reproductive development\citep{gerik2024sorghum}. The visualization demonstrates a structured sampling strategy that balances representation across genotypes, surfaces, and regions while maintaining sufficient sample sizes for each combination. The balanced spread across all dimensions confirms that the dataset successfully captures broad biological diversity.

This structured sampling approach provided a diverse image set, which allowed our dataset to reflect some of the natural variability in sorghum stomata and maintain the consistent image quality required for reliable annotation and downstream processing.

\begin{figure}[h]
  \centering
  \includegraphics[width=0.95\textwidth]{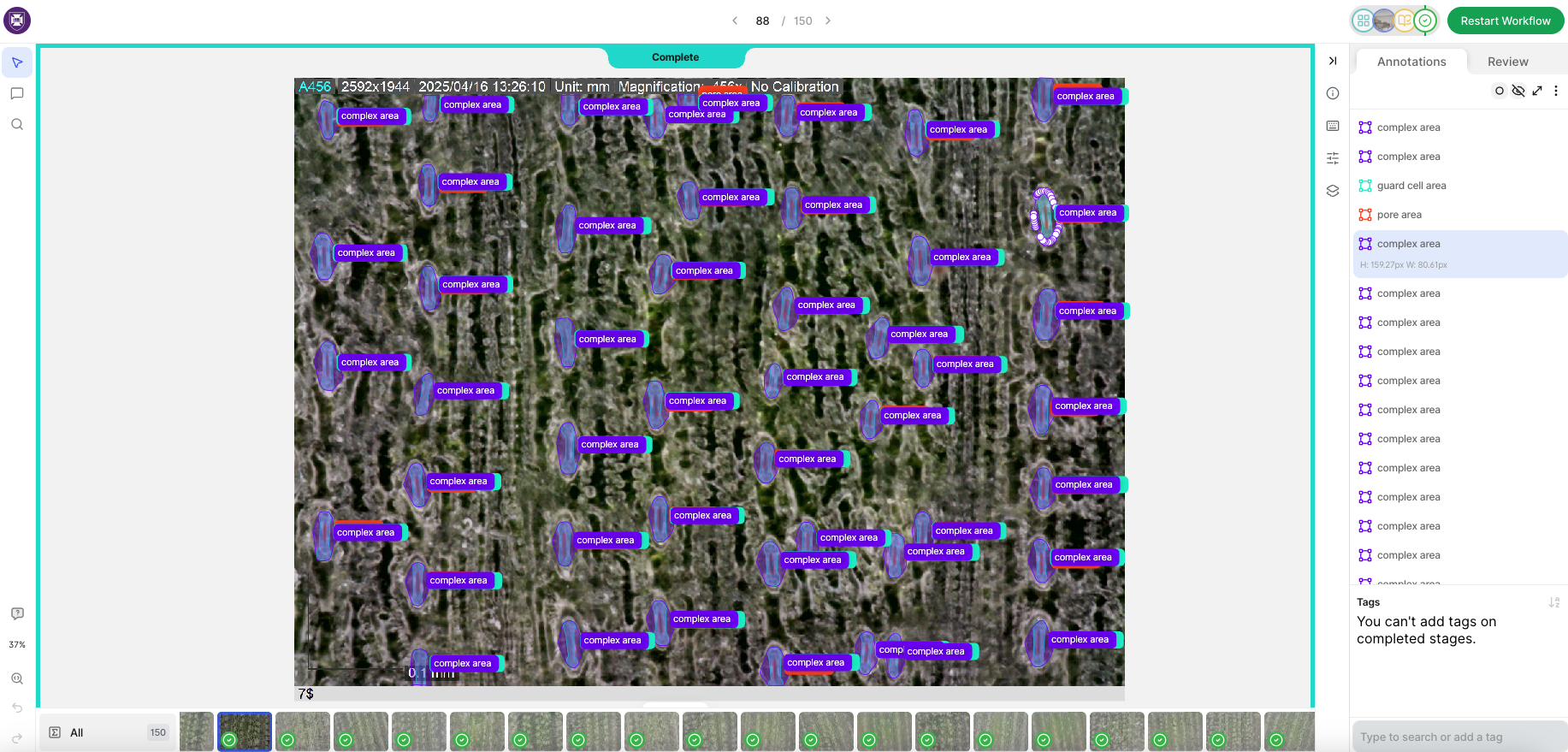}
  \caption{Annotation interface on the V7 platform showing pixel-wise labelling of a sorghum image. Each stomatal complex is delineated, with inner guard cell and pore areas annotated as distinct objects. Annotation classes are colour-coded and listed in the sidebar to support real-time collaboration, instance management, and quality monitoring. The interface also displays metadata and annotation progress to ensure consistent and efficient data curation.}
  \label{fig:v7_annotation}
\end{figure}

\subsection{Segmentation annotation}
\label{subsec:annotation}

The pixel-wise annotation of stomatal features was performed using the V7 platform \citep{V7Docs} (as shown in Fig.~\ref{fig:v7_annotation}), which is a cloud-based annotation tool selected for its advanced support of mask polygon annotation, collaborative workflows and quality tracking. Annotation was performed by three trained annotators with prior experience in stomatal trait analysis and followed a detailed guideline jointly developed with domain experts in stomata research. Prior to the main annotation phase, all annotators completed a training set and a consensus round to ensure a consistent interpretation of class boundaries and object definitions. Each annotated object was assigned to one of the three stomatal classes: complex area, guard cell area, or pore area.

Annotators were instructed to use the polygon tool to draw precise, closed contours for each object and minimise overlap and ambiguity at class boundaries. Each image typically contains dozens of stomatal complexes, with individual instances annotated for all three classes when clearly visible. All annotated masks are single-instance, non-overlapping, and pixel-precise, supporting downstream trait quantification and model training. Throughout the annotation process, class assignments were colour coded in the V7 interface, with object lists and metadata displayed in the same sidebar for transparency and tracking. Real-time collaboration features allowed annotators to flag uncertain cases, consult experts, and resolve discrepancies promptly. Quality tracking and progress were monitored within the tool to maximize performance and maintain consistency. Finally, the completed annotations were exported directly in COCO-style JSON formats. 

This rigorous annotation workflow ensured granularity and precision of class polygons, as well as consistency and traceability across all dataset images. Combined with expert review and iterative correction, these protocols underpin the high quality and research utility of the sorghum stomata dataset.

\begingroup
\setlength{\tabcolsep}{4pt}             
\renewcommand{\arraystretch}{0.9}        

\begin{table*}[t]
\centering
\caption{Annotation effort and quality-control summary of the dataset.}
\label{tab:annotation_overview}
\small                                    
\resizebox{0.9\textwidth}{!}{%
\begin{tabular}{c c c c}
\hline
\textbf{Annotations} & \textbf{Time per image} & \textbf{Time per annotation} & \textbf{Review pass rate} \\
\hline
40\,750 & 25 m 25 s & 15 s & 90\% \\
\hline
\end{tabular}%
}
\end{table*}

\endgroup

\subsection{Risk management of annotation quality}
\label{subsec:quality_control}

To ensure that our sorghum stomata dataset reached the highest standards of reliability and biological accuracy, a comprehensive quality control framework was implemented throughout the annotation pipeline. This framework comprised multiple stages of pre-annotation training, guideline development, systematic expert review, and iterative correction designed to minimize errors, eliminate ambiguity, and maintain consistency between annotators.

Before formal annotation began, a detailed set of annotation guidelines was collaboratively established by domain experts and project leaders. These guidelines covered class definitions, boundary conventions, example edge cases, and instructions for handling ambiguous or degraded image regions. All annotators completed a consensus training round on a subset of images, with discrepancies discussed and resolved in consultation with experts to ensure a unified interpretation of biological structures.

During the main annotation phase, every annotated image underwent a structured review process. Domain experts systematically checked each image and its corresponding masks for annotation errors such as missed or mislabelled stomata, incorrect class assignments, boundary inaccuracies, or mask overlaps. Any image found to contain errors was returned to the original annotator for correction, with additional feedback provided as necessary. This iterative review re-annotation loop was repeated until each image met strict quality criteria.

Progress and performance were continuously monitored using the built-in tracking features and external logs. The summary of the dataset shown in Table~\ref{tab:annotation_overview} reflects a significant manual investment. The achieved 90\% review pass rate is a direct indicator of both annotator diligence and the rigour of the quality assurance workflow. The median annotation time per image and per object further demonstrate the sustained attention to detail required for accurate stomatal segmentation at scale.

To further ensure dataset integrity, regular group meetings were held throughout the project to discuss challenging cases, refine guidelines, and ensure consistent application of quality standards across annotators and time. The final dataset therefore provides confidence in its suitability for downstream analysis and benchmarking.

\subsection{Patch-based preprocessing}
\label{subsec:preprocessing}

To address the challenge of detecting small stomatal structures in high-resolution ($2592 \times 1944$) microscopy images, we employed a patch-based preprocessing strategy. Each raw image was cropped into $341 \times 341$ pixel patches with a 10-pixel overlap, using a stride of 331 pixels along both horizontal and vertical axes. Additional tiles were appended near the image borders to ensure full spatial coverage of the original frame.

Although we took care to capture clear images, some leaf regions still contained blur or slight defocus due to surface curvature. To avoid incorrect human annotations in such cases, annotators were instructed not to label any stomatal structure unless it was clearly visible. To enforce this in the preprocessing stage for the human-annotated subset, we applied two filtering criteria: (1) a stomatal instance was assigned to a patch only when more than 50\% of its area lay within that patch, preventing duplicate counting across overlaps; and (2) patches without any human stomatal annotation were excluded, effectively removing blurred, out-of-focus, or ambiguous regions from the manually annotated set.

After filtering, each retained patch contained on average three stomata, increasing the apparent size of fine structures such as pores and guard cells relative to the patch dimensions. This produced 7,662 training patches from 222 original images, 2,238 validation patches from 63 images, and 1,160 test patches from 33 images, while approximately 25\% of the total were empty patches that were excluded from the human-annotated subset. The resulting patch set substantially enhances small-object representation and ensures more balanced spatial sampling across the dataset.

\subsection{Pseudo labels generation and semi-supervised expansion}
\label{subsec:pseudo_label_generation}

To extend the effective training set beyond the 11,060 manually annotated patch images, a semi-supervised pseudo-labelling framework was developed using the Mask R-CNN model \citep{he2017mask} with a ConvNeXt-V2-Base backbone \citep{Woo2023ConvNeXtV2}. 
In addition to unlabelled images collected within our own imaging pipeline, we incorporated 80 sorghum microscopy images provided by Prof Krishna Jagadish into the unlabelled image pool. These images exhibit similar stomatal appearance, resolution, and contrast characteristics to our collected images. This allowed the seed model to generate accurate pseudo labels without additional domain adaptation. At the same time, they were captured under different experimental settings, which introduces a complementary variation in the imaging conditions and leaf presentation. Including these images therefore expands the diversity of the unlabelled dataset while remaining compatible with the learned feature representations. 

\begin{figure*}[t]
  \centering
  \includegraphics[width=0.95\textwidth]{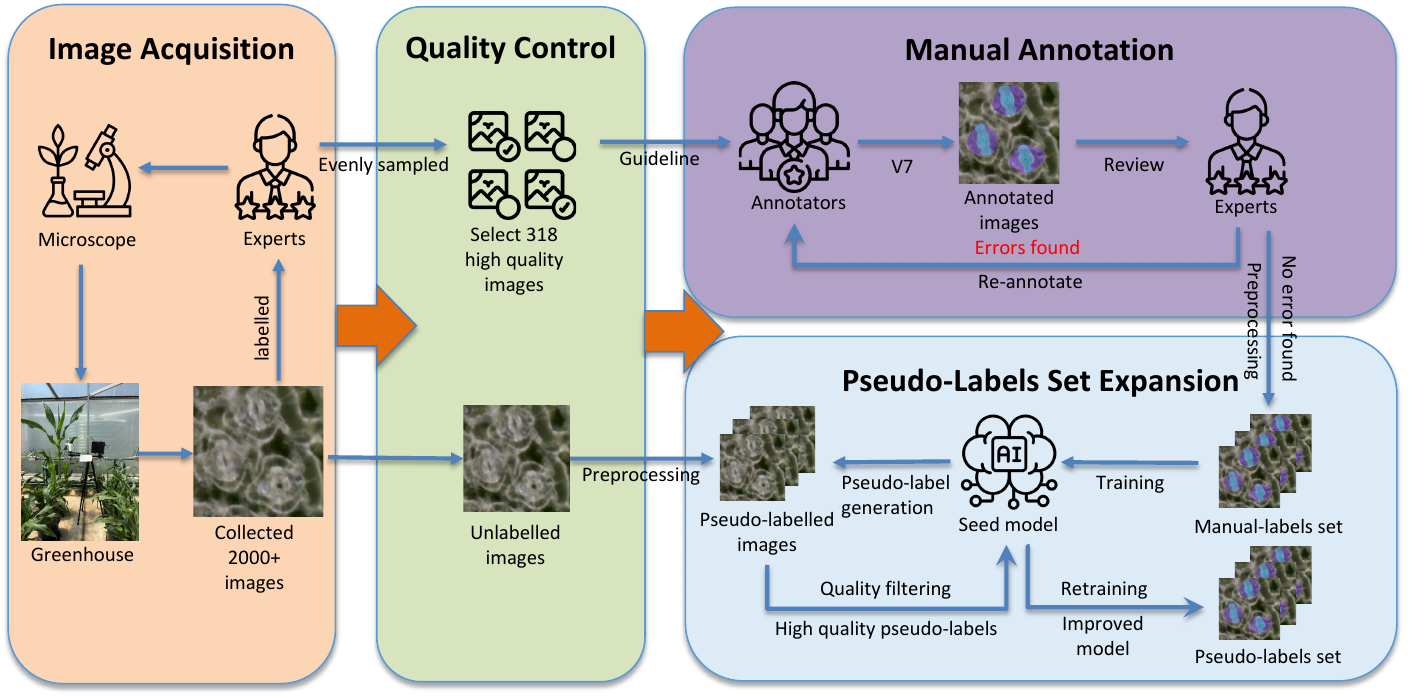}
  \caption{Dataset curation and expansion workflow. The pipeline comprises: (1) image acquisition was performed inside a solarweave greenhouse with microscopy; (2) quality filtering of the 318 original images ensuring balanced genotype, surface and region coverage; (3) manual pixel-wise annotation by trained annotators with expert review; and (4) pseudo-labelling expansion using only high-confidence predictions from the seed segmentation model.}
  \label{fig:workflow}
\end{figure*}

Following the same patch-based preprocessing strategy, all unlabelled images were divided into $341 \times 341$ pixel patches with a 10-pixel overlap to preserve stomatal structures near the patch boundaries. As illustrated in Fig.~\ref{fig:workflow}, an initial seed model trained on the manually annotated core set was employed to predict instance masks on the unlabelled patches. Each detected stomatal class received an individual confidence score, which enabled selective instance-level filtering. This instance-based approach was intentionally chosen for pseudo-labelling because instance segmentation provides confidence scores at the instance level rather than pixel level, as in semantic segmentation. Consequently, entire high-quality instance masks can be retained while uncertain or incomplete detections are excluded, producing coherent pseudo labels with accurate boundaries and lower fragmentation. This instance-wise filtering mechanism is critical for controlling label noise and maintaining the integrity of automatically generated annotations during model retraining.

To enhance pseudo labels reliability, class-specific confidence thresholds were applied: 0.5 for the pore area, 0.7 for the guard cell area, and 0.7 for the complex area. Patches without confident detections were also preserved to maintain a balanced distribution of stomata and background, which is essential for stable semi-supervised learning. 

This pseudo-labelling and retraining process significantly improved model generalisation by exploiting unlabelled data while constraining noise propagation, and it also provides a scalable method for refining segmentation performance under limited manual annotation budgets.

\subsection{Segmentation model and evaluation metrics}
\label{subsec:model_evaluation}

To establish reliable baselines and assess the benefits of patch-based preprocessing and pseudo-label expansion, we conducted systematic experiments using both semantic and instance segmentation frameworks. All models were trained and evaluated on the StomataSeg dataset, which includes human-annotated and pseudo-labelled images divided into training, validation, and test splits as described in Section~\ref{subsec:preprocessing}.

\subsubsection{Model architectures}

For semantic segmentation, we evaluated six representative models from the MMSegmentation framework. These models cover both classical convolutional designs and more recent transformer-based architectures. The set includes Fully Convolutional Network (FCN; \citealp{long2015fully}), Object-Contextual Representations (OCRNet; \citealp{yuan2020object}), BiSeNetV2 \citep{yu2021bisenet}, UPerNet \citep{xiao2018unified}, SegNeXt \citep{guo2022segnext}, and SegFormer \citep{xie2021segformer}. Each model was trained with its recommended backbone to ensure consistency: ResNet variants \citep{resnet} for FCN and OCRNet, MSCAN \citep{guo2022segnext} for SegNeXt, Mix Transformer (MIT) \citep{xie2021segformer} for SegFormer, Twins-SVT-S \citep{chu2021twins} for UPerNet, and BiSeNetV2’s own lightweight architecture \citep{yu2021bisenet}. All models were trained on patch images using four semantic classes, including background.

For instance segmentation, we used six widely adopted models from the MMDetection framework. Most are based on the Mask R-CNN architecture \citep{he2017mask}, paired with different backbones to examine the role of feature extractors. These include ResNet-50 \citep{resnet}, Swin Transformer Tiny (Swin-T; \citealp{liu2021Swin}), Pyramid Vision Transformer v2-B2 (PVT v2-B2; \citealp{wang2021pvtv2}), and ConvNeXt-V2-Base \citep{Woo2023ConvNeXtV2}. We also included Mask2Former \citep{cheng2021mask2former} and Hybrid Task Cascade (HTC; \citealp{chen2019hybrid}) with ResNet-50 backbones, representing alternative instance-segmentation approaches. All instance models were trained on three stomatal classes for 500 epochs using the AdamW optimizer. Learning rates and batch sizes were adjusted for each backbone to maintain stable training and efficient GPU usage.

\subsubsection{Dataset splits}
The original 318 images were divided into 222, 63, and 33 images for training, validation, and testing, respectively. These correspond to 7,662 training patches, 2,238 validation patches, and 1,160 test patches. For pseudo-label expansion experiments, an additional 56,428 automatically labelled patches were integrated into training, combining with 7,662 human-annotated patches to form the semi-supervised training set of 64,090 images.

\begin{figure*}[t]
  \centering
  \includegraphics[width=0.85\textwidth]{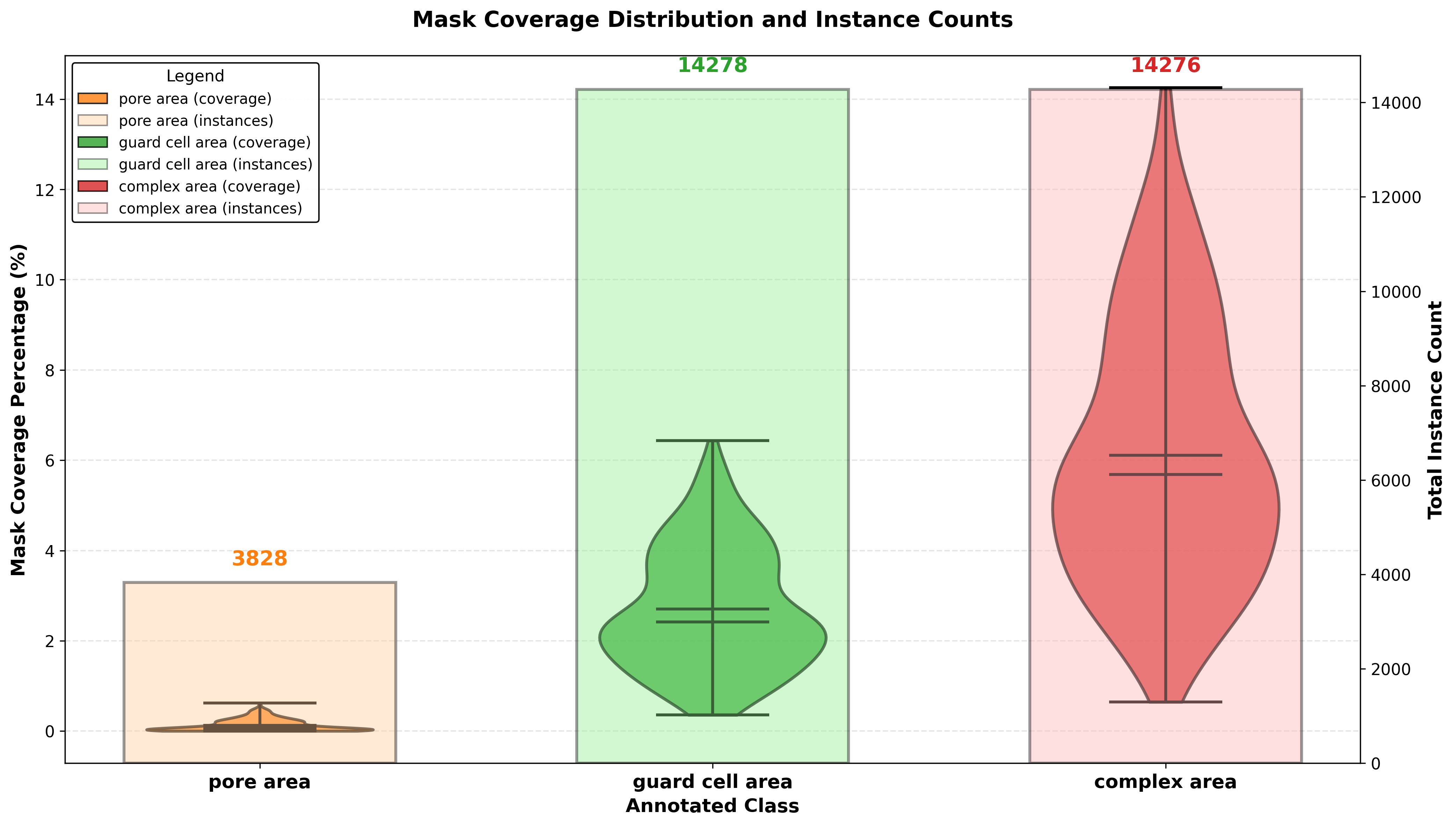}
  \caption{Mask coverage distribution and instance-count summary for each annotated class in the dataset. Violin plots (foreground, darker colours) display kernel-density distributions of mask coverage percentages per class, with internal lines showing medians and quartiles. The bar charts (background, lighter colours) present total annotated instance counts. The stomatal complex and guard cell classes show similar distributions and frequencies, while the pore class exhibits much lower coverage and roughly one-quarter the instance count, reflecting its smaller geometry and reduced detection frequency due to aperture-state variation.}
  \label{fig:mask_distribution}
\end{figure*}

\subsubsection{Evaluation metrics}
Semantic segmentation models were assessed using mean Intersection over Union (mIoU) and mean Accuracy (mAcc). The mIoU metric captures average overlap between predicted and ground-truth regions across all classes, while mAcc reflects the average per-class pixel-level accuracy:
\begin{align}
\text{mIoU} = \frac{1}{C} \sum_{c=1}^{C} \frac{TP_c}{TP_c + FP_c + FN_c}, \qquad
\text{mAcc} = \frac{1}{C} \sum_{c=1}^{C}\frac{TP_c}{TP_c + FN_c},
\end{align}
where $C$ is the number of classes, and $TP_c$, $FP_c$, and $FN_c$ denote true positive, false positive, and false negative pixel counts for class $c$, respectively.

Instance segmentation performance was evaluated using mean Average Precision (mAP) across IoU thresholds from 0.5 to 0.95 in increments of 0.05, denoted as AP, and at IoU threshold 0.5 (AP$_{50}$). We also report per-class AP for the pore, guard cell, and complex area classes:
\begin{equation}
\text{mAP} = \frac{1}{N} \sum_{i=1}^{N} \text{AP}_i, \qquad
\text{AP}_i = \int_0^1 p(r) \, dr,
\end{equation}
where $N$ is the number of classes and $p(r)$ represents the precision–recall curve for class $i$.

Together, these metrics provide a comprehensive assessment of both pixel-level and instance-level segmentation accuracy, which allow fair comparison across architectures and training regimes.

\begin{figure}[t]
  \centering
  \includegraphics[width=0.85\textwidth]{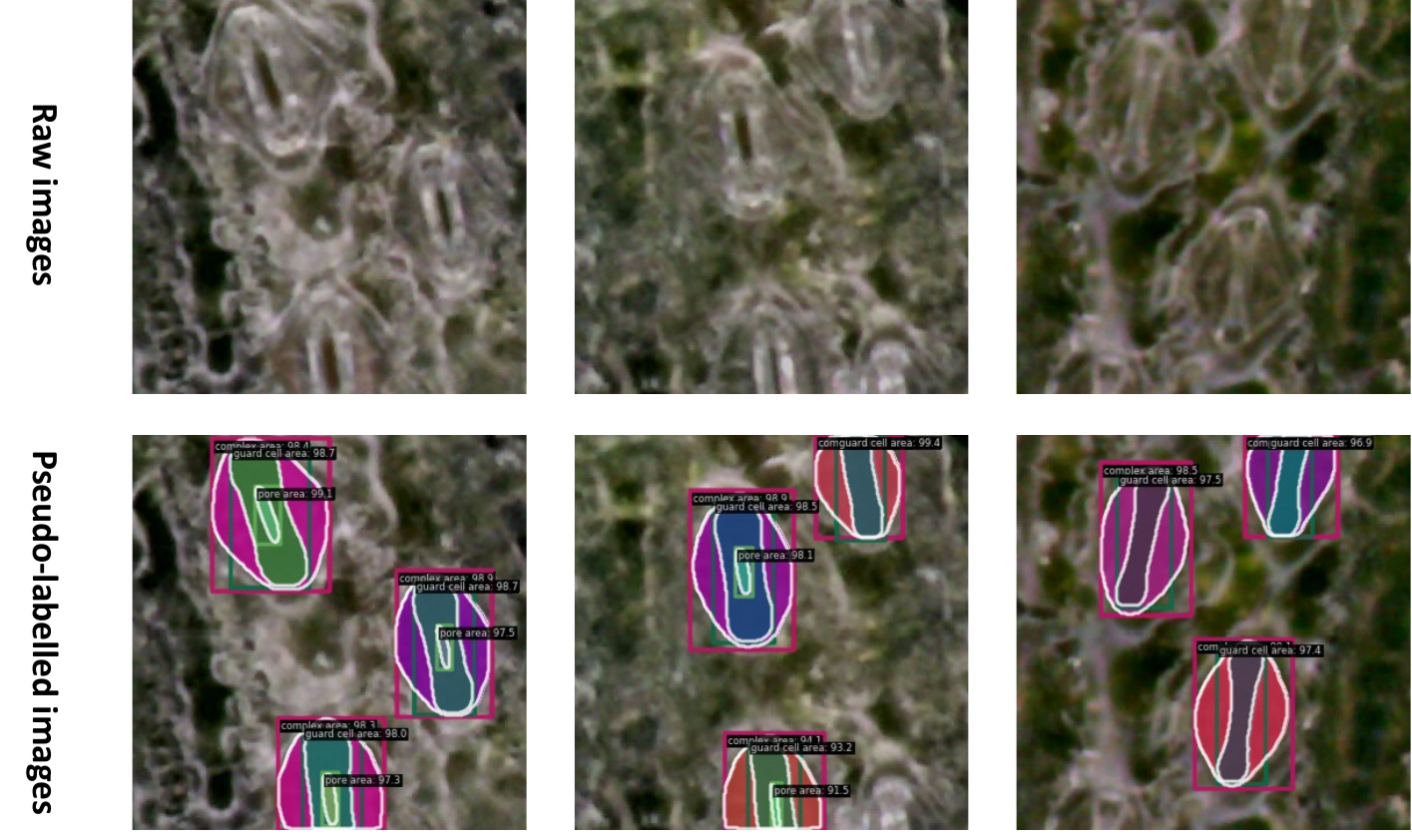}
  \caption{Illustration of high-confidence pseudo labels used in our semi-supervised segmentation framework. Example patches show instance masks for stomatal complex, guard cell and pore areas annotated by the seed model and retained through confidence-based filtering. These pseudo-labelled samples supplement the manually annotated core set, enabling our pipeline to expand training coverage and improve model generalisation across variable stomatal appearances.}
  \label{fig:pseudo_labels}
\end{figure}

\begin{table*}[t]
\centering
\small
\caption{Benchmarking results of semantic segmentation models using original full-resolution images and patch-based preprocessing. Results include overall mIoU/mAcc and per-class mIoU/mAcc.}
\vspace{-10pt}
\label{tab:semantic_segmentation}
\resizebox{\textwidth}{!}{%
\begin{tabular}{llcccccccc}
\toprule
\textbf{Method} & \textbf{Backbone} &
\multicolumn{2}{c}{\textbf{Overall}} &
\multicolumn{2}{c}{\textbf{Pore area}} &
\multicolumn{2}{c}{\textbf{Guard cell}} &
\multicolumn{2}{c}{\textbf{Complex area}} \\
\cmidrule(lr){3-4} \cmidrule(lr){5-6} \cmidrule(lr){7-8} \cmidrule(lr){9-10}
 & & \textbf{mIoU} & \textbf{mAcc} &
\textbf{mIoU} & \textbf{mAcc} &
\textbf{mIoU} & \textbf{mAcc} &
\textbf{mIoU} & \textbf{mAcc} \\
\midrule
\multicolumn{10}{l}{\textbf{Original images}} \\
\midrule
OCRNet \cite{yuan2020object} & ResNet-50 & 49.21 & 58.86 & 13.83 & 16.01 & 48.82 & 69.00 & 36.89 & 51.77 \\
BiSeNetV2 \cite{yu2021bisenet} & BiSeNetV2 & 49.73 & 57.52 & 2.18 & 2.22 & 54.33 & 69.30 & 44.88 & 59.58 \\
FCN \cite{long2015fully} & ResNet-18 & 52.14 & 60.59 & 5.95 & 6.16 & 57.87 & 75.93 & 46.96 & 61.33 \\
FCN \cite{long2015fully} & ResNet-50 & 52.48 & 61.35 & 6.00 & 6.16 & 58.87 & 77.75 & 47.27 & 62.62 \\
SegNeXt \cite{guo2022segnext} & MSCAN-T & 61.49 & 71.43 & 32.12 & 38.10 & 64.36 & 80.92 & 51.56 & 67.81 \\
SegNeXt \cite{guo2022segnext} & MSCAN-S & 62.04 & 72.10 & 32.09 & 38.36 & 65.28 & 82.04 & 52.81 & 69.11 \\
UPerNet \cite{xiao2018unified} & Twins-SVT-S & 61.57 & 72.40 & 36.09 & 44.00 & 62.99 & 79.01 & 49.51 & 67.91 \\
SegFormer \cite{xie2021segformer} & MIT-B0 & 65.09 & 75.21 & 48.33 & 57.38 & 63.78 & 78.24 & 50.45 & 66.30 \\
SegFormer \cite{xie2021segformer} & MIT-B1 & 65.93 & 77.39 & 48.99 & 61.97 & 64.93 & 81.51 & 51.91 & 67.20 \\
\midrule
\multicolumn{10}{l}{\textbf{Patches}} \\
\midrule
UPerNet \cite{xiao2018unified} & Twins-SVT-S & 62.81 & 73.87 & 50.49 & 60.38 & 58.31 & 76.73 & 45.41 & 59.80 \\
SegNeXt \cite{guo2022segnext} & MSCAN-S & 63.43 & 75.90 & 50.75 & 62.69 & 60.02 & 79.33 & 45.98 & 63.32 \\
BiSeNetV2 \cite{yu2021bisenet} & BiSeNetV2 & 67.20 & 79.63 & 55.09 & 73.41 & 64.38 & 79.81 & 51.97 & 66.61 \\
FCN \cite{long2015fully} & ResNet-50 & 67.32 & 80.27 & 52.47 & 72.24 & 65.72 & 82.43 & 53.62 & 67.71 \\
OCRNet \cite{yuan2020object} & ResNet-50 & 67.33 & 79.68 & 51.39 & 66.27 & 66.00 & 84.76 & 54.38 & 69.10 \\
SegFormer \cite{xie2021segformer} & MIT-B1 & \textbf{70.35} & \textbf{81.97} & \textbf{60.26} & \textbf{76.92} & \textbf{68.03} & \textbf{83.58} & \textbf{55.47} & \textbf{68.51} \\
\bottomrule
\end{tabular}%
}
\end{table*}

\begin{table*}[t]
\centering
\small
\caption{Benchmarking results of instance segmentation models using original full-resolution images and patch-based preprocessing. Results include overall AP/AP$_{50}$ and per-class AP/AP$_{50}$.}
\vspace{-10pt}
\label{tab:instance_segmentation}
\resizebox{\textwidth}{!}{%
\begin{tabular}{llcccccccc}
\toprule
\textbf{Method} & \textbf{Backbone} &
\multicolumn{2}{c}{\textbf{Overall}} &
\multicolumn{2}{c}{\textbf{Pore area}} &
\multicolumn{2}{c}{\textbf{Guard cell}} &
\multicolumn{2}{c}{\textbf{Complex area}} \\
\cmidrule(lr){3-4} \cmidrule(lr){5-6} \cmidrule(lr){7-8} \cmidrule(lr){9-10}
 & & \textbf{AP} & \textbf{AP$_{50}$} &
\textbf{AP} & \textbf{AP$_{50}$} &
\textbf{AP} & \textbf{AP$_{50}$} &
\textbf{AP} & \textbf{AP$_{50}$} \\
\midrule
\multicolumn{10}{l}{\textbf{Original images}} \\
\midrule
Mask R-CNN \cite{he2017mask} & ResNet-50 & 26.17 & 50.61 & 0.00 & 0.00 & 31.84 & 68.60 & 46.67 & 83.23 \\
Mask2Former \cite{cheng2021mask2former} & ResNet-50 & 25.37 & 46.36 & 0.29 & 0.99 & 29.47 & 60.46 & 46.36 & 77.63 \\
HTC \cite{chen2019hybrid} & ResNet-50 & 26.36 & 51.15 & 0.00 & 0.00 & 35.56 & 76.27 & 43.54 & 77.16 \\
Mask R-CNN \cite{he2017mask} & Swin-T & 26.00 & 51.10 & 0.00 & 0.00 & 27.42 & 66.03 & 50.63 & 87.41 \\
Mask R-CNN \cite{he2017mask} & PVTv2-B2 & 26.90 & 50.90 & 0.00 & 0.00 & 31.86 & 67.92 & 48.90 & 84.72 \\
Mask R-CNN \cite{he2017mask} & ConvNeXt-V2-Base & 28.30 & 50.70 & 0.59 & 0.99 & 33.91 & 67.96 & 50.52 & 83.12 \\
\midrule
\multicolumn{10}{l}{\textbf{Patches}} \\
\midrule
Mask2Former \cite{cheng2021mask2former} & ResNet-50 & 35.80 & 74.60 & 22.40 & 61.20 & 38.80 & 81.50 & 46.20 & 81.10 \\
HTC \cite{chen2019hybrid} & ResNet-50 & 42.00 & 79.70 & 32.30 & 76.00 & 42.70 & 81.00 & 51.00 & 82.00 \\
Mask R-CNN \cite{he2017mask} & ResNet-50 & 43.80 & 84.70 & 32.20 & 78.20 & 45.20 & 86.80 & 54.00 & 89.10 \\
Mask R-CNN \cite{he2017mask} & Swin-T & 43.90 & 83.50 & 35.40 & 77.80 & 45.00 & 87.20 & 51.30 & 85.30 \\
Mask R-CNN \cite{he2017mask} & PVTv2-B2 & 44.10 & 83.00 & 33.90 & 71.60 & 44.70 & 88.50 & 53.70 & 89.00 \\
Mask R-CNN \cite{he2017mask} & ConvNeXt-V2-Base & \textbf{46.10} & \textbf{87.60} & \textbf{33.10} & \textbf{82.70} & \textbf{48.90} & \textbf{89.90} & \textbf{56.30} & \textbf{90.10} \\
\bottomrule
\end{tabular}}
\end{table*}

\section{Results and discussion}
\label{sec:results}

\subsection{Dataset insights and training implications}
\label{subsec:data_characteristics}

The StomataSeg dataset assembled in this study consists of 67,488 patch images, including 11,060 human-annotated samples and 56,428 pseudo-labelled samples. Whereas Section~\ref{sec:materials_methods} described the acquisition and labelling workflow in detail, this subsection highlights the dataset characteristics relevant for model training and interpretation.

The annotated images have a clear imbalance among the three stomatal component areas (Fig.~\ref{fig:mask_distribution}). The stomatal complex area has the largest and most variable spatial coverage and also appears more frequently. In contrast, the pore area accounts for only about one-quarter as many instances due to its small size and its inconsistent visibility in microscopy images. This creates a typical small-object segmentation challenge, as the pores occupy only a small portion of each patch and may be partially occluded or closed. This creates a typical small-object segmentation challenge. The guard cell area falls between these two extremes, as it showed moderate coverage.

In addition to the manually annotated subset, our semi-supervised pseudo-labelling pipeline generated 56,428 supplementary patches, expanding the dataset by more than fivefold without additional human annotation effort. Pseudo labels (examples shown in Fig.~\ref{fig:pseudo_labels}) are derived from high-confidence model predictions on unlabelled patches. Given that the seed model has already been trained on diverse annotated data, it can reliably predict the absence of stomata. We filtered pseudo labels using class-specific confidence thresholds that were selected through manual validation to ensure that retained detections are highly reliable. Under these thresholds, patches without confident detections are still retained in the pseudo-labelled set. This is distinct from the practice for our human-annotated subset, where empty patches were removed to avoid including unclear or blurred regions that annotators were instructed to skip. We retained genuine empty patches during the pseudo-labelling process to strengthen background discrimination during training. This practice helps reduce false positives and further improves model robustness across both stomatal and non-stomatal regions.

The dataset analysis highlights two main implications for model development. Firstly, patch-based preprocessing plays an important role in addressing the strong scale disparity between stomatal components and background, allowing the model to capture fine structures such as pores with higher spatial resolution. Secondly, the pseudo-labelling process also expands the dataset to cover a wider range of genotypes, leaf surfaces, and imaging conditions. This increases its representativeness and supports effective semi-supervised learning for biologically diverse segmentation tasks.

\subsection{Effect of segmentation performance}
\label{subsec:benchmarking_phenotyping}

We conducted comprehensive benchmarking across both semantic and instance segmentation frameworks in order to assess the impact of our patch-based preprocessing and semi-supervised expansion strategy on stomatal component segmentation, and to explore implications for downstream phenotyping. To evaluate the effect of patch-based preprocessing, we compared model performance when trained on full-frame high-resolution images versus patch-based images. Table~\ref{tab:semantic_segmentation} and Table~\ref{tab:instance_segmentation} summarize the results for the representative semantic and instance segmentation architectures in both regimes.

\begin{figure*}[t]
    \centering
    \includegraphics[width=0.95\textwidth]{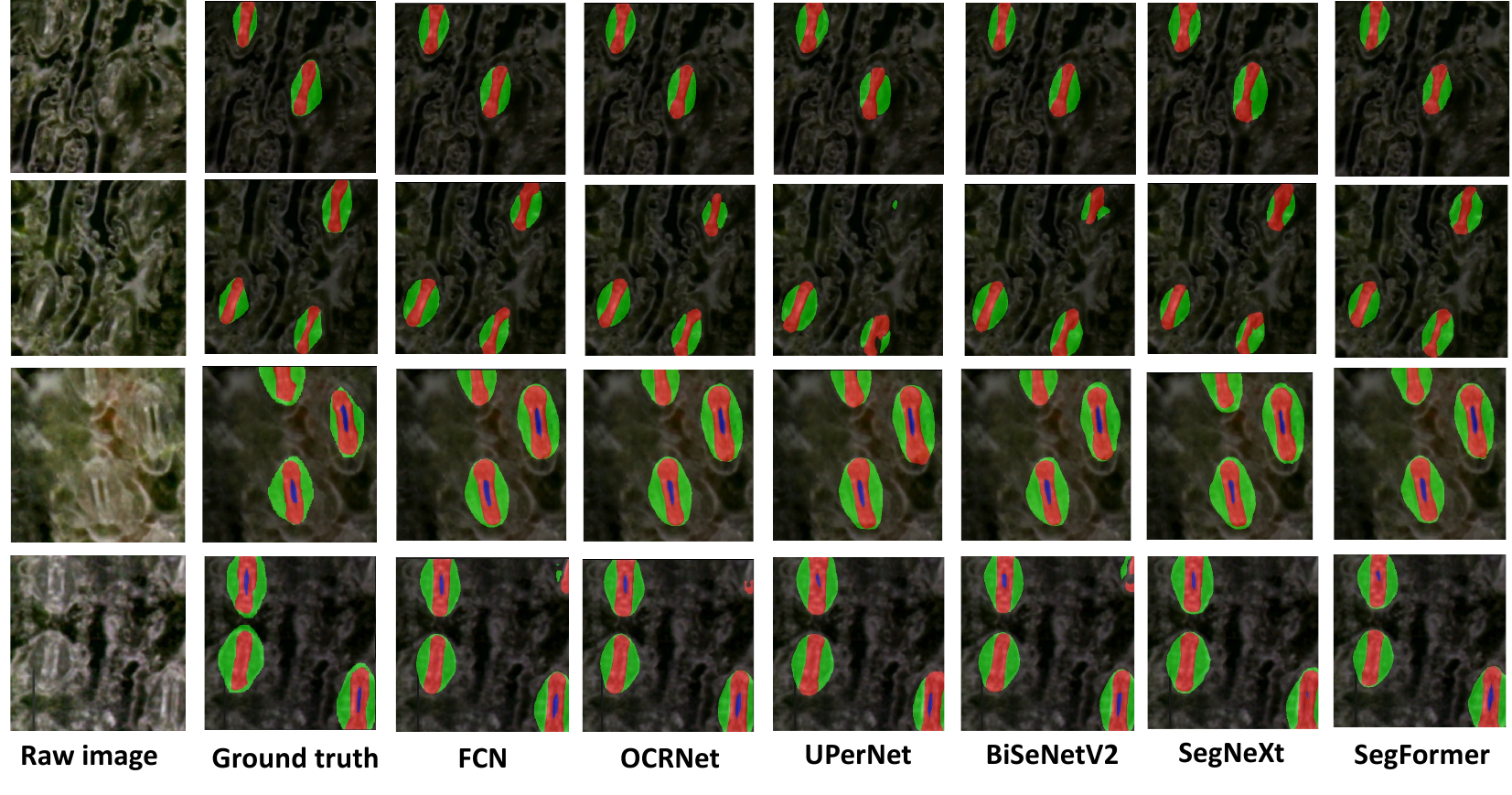}
    \caption{\textit{Visualization of semantic segmentation results on stomata images.} Green mask shows complex area, red mask shows guard cell area and blue mask shows pore area. From left to right: raw image, ground-truth masks, predicted segmentation masks from different methods.}
    \vspace{-10pt}
    \label{fig:semantic_seg}
\end{figure*}

\begin{figure*}[t]
    \centering
    \includegraphics[width=0.95\textwidth]{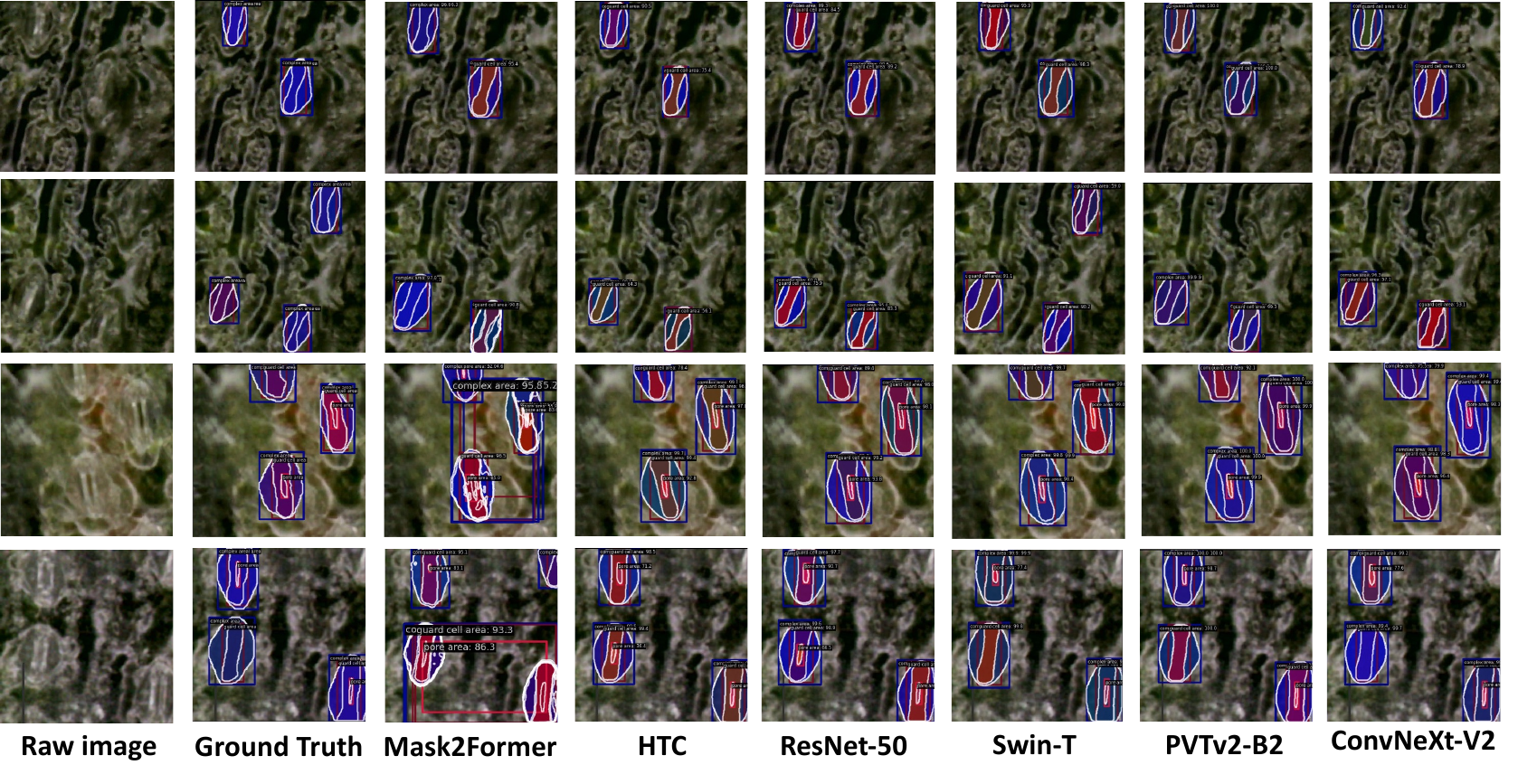}
    \vspace{-10pt}
    \caption{\textit{Visualization of instance segmentation results on stomata images.} From left to right: raw image, ground-truth masks, predicted segmentation masks from different methods.}
    \label{fig:instance_seg}
\end{figure*}

The results reveal two key findings. First, segmentation based on original full-frame images remains highly challenging, especially for the pore area, where most semantic segmentation and all instance segmentation methods fail to segment it out. Second, our patch-based preprocessing strategy yields substantial improvements compared to using the original images: for semantic models the top mIoU increases from 65.93\% to 70.35\% (Table~\ref{tab:semantic_segmentation}), whereas for instance models the top AP rises from 28.30\% to 46.10\% (Table~\ref{tab:instance_segmentation}). Notably, pore area AP improves from near zero to 35.40\%, transforming what was essentially a failure case into a promising segmentation performance (as shown in Fig.~\ref{fig:semantic_seg} and Fig.~\ref{fig:instance_seg}).

\begin{table*}[t]
\centering
\caption{Performance comparison of Mask R-CNN with ConvNeXt-V2 before and after pseudo-label retraining.}
\label{tab:retrain_results}
\small
\resizebox{\textwidth}{!}{%
\begin{tabular}{l l cc cc cc cc}
\toprule
\textbf{Method} & \textbf{Training Data} &
\multicolumn{2}{c}{\textbf{Overall}} &
\multicolumn{2}{c}{\textbf{Pore Area}} &
\multicolumn{2}{c}{\textbf{Guard Cell}} &
\multicolumn{2}{c}{\textbf{Complex Area}} \\
\cmidrule(lr){3-4} \cmidrule(lr){5-6} \cmidrule(lr){7-8} \cmidrule(lr){9-10}
 & & \textbf{AP} & \textbf{AP$_{50}$} & \textbf{AP} & \textbf{AP$_{50}$} & \textbf{AP} & \textbf{AP$_{50}$} & \textbf{AP} & \textbf{AP$_{50}$} \\
\midrule
Seed Model & 7,662 GT patches & 46.10 & 87.60 & 33.10 & 82.70 & 48.90 & 89.90 & 56.30 & 90.10 \\
After Retraining & 7,662 GT + 56,428 PL & \textbf{49.20} & \textbf{88.80} & \textbf{39.10} & \textbf{82.90} & \textbf{50.30} & \textbf{90.60} & \textbf{58.30} & \textbf{93.10} \\
Improvement & -- & \textbf{+3.10} & \textbf{+1.20} & \textbf{+6.00} & \textbf{+0.20} & \textbf{+1.40} & \textbf{+0.70} & \textbf{+2.00} & \textbf{+3.00} \\
\bottomrule
\end{tabular}%
}
\end{table*}

\begin{figure*}[hp]
  \centering
  \includegraphics[width=0.9\textwidth]{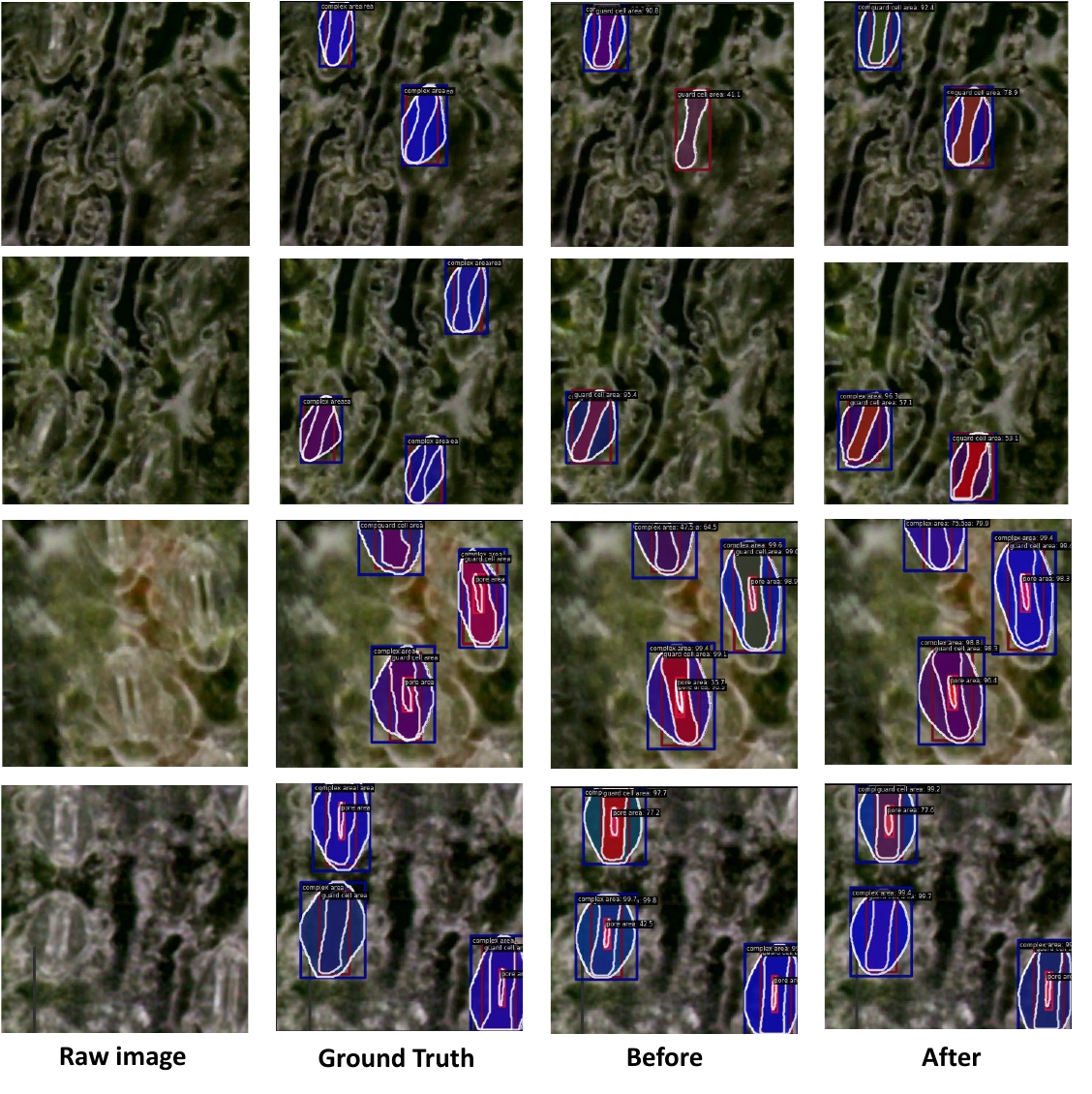}
  \caption{Visual comparisons of segmentation results before and after pseudo-label retraining using the Mask R-CNN with ConvNeXt-V2 backbone. Each row shows the same stomatal image with (a) raw image, (b) ground-truth annotations, (c) predictions before pseudo-labelling, and (d) predictions after pseudo-labelling.}
  \label{fig:pseudo_label_retrain}
\end{figure*}

To further evaluate the effectiveness of the proposed semi-supervised training framework, we retrained the leading instance segmentation model by combining 7,662 human-annotated patches with 56,428 high-confidence pseudo-labelled patches. Table~\ref{tab:retrain_results} summarises the performance improvements achieved after retraining, while Fig.~\ref{fig:pseudo_label_retrain} provides qualitative examples of segmentation results before and after pseudo-label retraining.

The results show that the framework uses unlabelled data effectively to improve model robustness and generalisation. Pseudo labels filtered by class-specific confidence thresholds increase data diversity while keeping noise low. The 6\% gain in average precision for the pore area suggests that the model learns finer morphological details of small and low-contrast structures that are often under-represented in manually annotated datasets.

Instance-level pseudo-labelling further strengthens this effect. Because confidence is assigned to each object rather than to individual pixels, the method preserves complete high-quality masks and removes uncertain detections. When combined with patch-based preprocessing, which enlarges local features and reduces resolution imbalance, the framework forms a scalable and interpretable pipeline that lowers manual annotation requirements and improves segmentation accuracy across all stomatal components.
 
\subsection{Foundation Model Baselines: SAM2.1 Zero-Shot Segmentation}
\label{subsec:sam2_baselines}

\begin{figure}[hp]
    \centering
\includegraphics[width=0.95\textwidth]{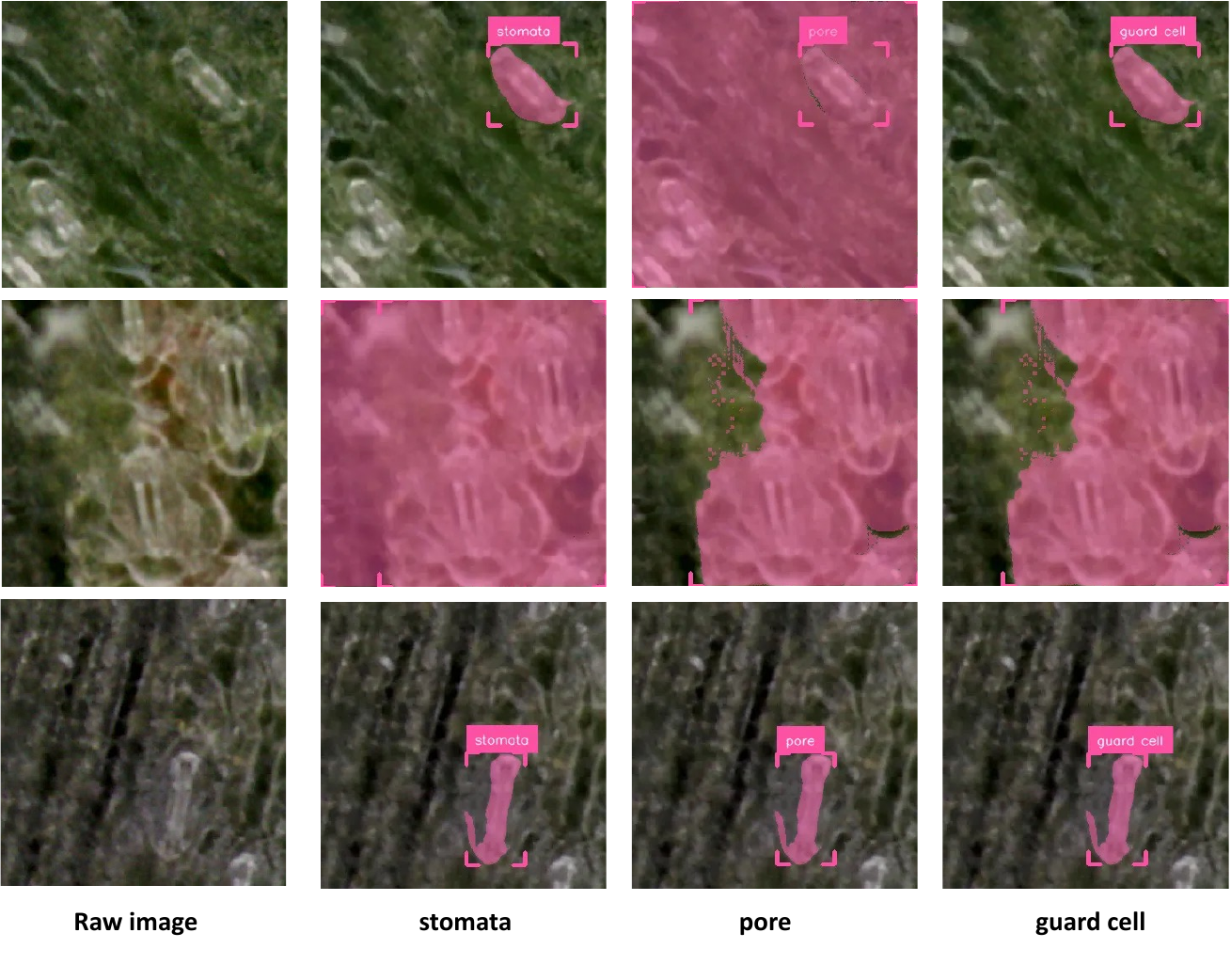}
    \caption{SAM2.1 zero-shot segmentation results on representative stomatal images using text prompts.}
    \label{fig:sam2_baselines}
\end{figure}

To contextualize the performance of our proposed pipeline, we evaluated SAM2.1's~\citep{ravi2024sam} open-vocabulary capabilities using text prompts through the \texttt{lang-segment-anything} framework\footnote{\url{https://github.com/luca-medeiros/lang-segment-anything}}, which combines GroundingDINO~\citep{liu2024grounding} for text-to-bounding-box detection with SAM2.1 for segmentation. As illustrated in Figure~\ref{fig:sam2_baselines}, our evaluation revealed significant limitations: the text-prompted approach fails to segment anything meaningful for most images, often producing masks that cover most region of the image. For the few images in which segmentation was attempted, different text prompts (e.g., ``stomata'', ``pore'', ``guard cell'') consistently produced identical bounding boxes and segmentation masks, regardless of the specific prompt used, indicating that the existing open-vocabulary detection pipeline fails to distinguish between the specific stomatal components.

These observations highlight the limitations of general-purpose foundation models for specialised multi-class segmentation tasks. Although text-prompted approaches are promising for open-vocabulary segmentation, they cannot distinguish closely related, hierarchically nested structures without explicit training on domain-specific data. In contrast, our proposed framework provides reliable multi-class segmentation of all three stomatal classes without struggling with ambiguous text prompts, which shows the value of domain-specific training strategies for specialised segmentation tasks.

\subsection{Limitations and perspective}
\label{subsec:limitations_perspective}

This study advances multi-class stomatal segmentation through patch-based preprocessing and semi-supervised pseudo-labelling framework, yet several limitations remain. The dataset comes from a single species, so the generalizability to other species, surfaces, or imaging conditions remains to be evaluated. Pseudo-labels expand the dataset but can still introduce noise or systematic bias in difficult cases, and this may affect model stability when applied to unseen data. Patch-based processing improves resolution for small structures, but the scale gap between stomatal components persists, and pore areas continue to show lower detection and segmentation reliability than larger structures. 

Future work should broaden the framework to test images of stomata with open pores under plentiful light, multiple genotypes and species, and imaging modalities through domain adaptation or cross-domain few-shot semantic segmentation methods like PATNet\citep{lei2022cross}. Self-supervised learning offers a path toward reducing dependence on dense annotations, and architectures designed specifically for nested micro-scale structures may improve pore segmentation performance. Large vision language models \citep{zhang2024vision} and domain-specific fine-tuning strategies such as VLSM-Adapter \citep{dhakal2024vlsm} also present promising avenues for specialised stomatal segmentation tasks, as they may combine strong semantic understanding with effective transfer learning. Finally, integrating segmentation outputs with downstream phenotyping modules, such as pore aperture dynamics or stomatal conductance estimation, will help connect computer vision pipelines with physiological and agronomic applications.

\section{Conclusion}
\label{sec:conclusion}

In this work, we developed a novel segmentation pipeline for stomatal imagery that addresses key challenges of pore detection, class imbalance and annotation cost. By combining patch-based preprocessing with a semi-supervised pseudo-labelling expansion of training data, we achieved marked improvements, compared with the baseline model trained only on manually annotated patches, in multi-class instance segmentation of all three stomatal components, including an approximately 6\% increase in pore area average precision. Benchmarking results show that our approach substantially outperforms conventional full-frame image based methods and offers a feasible path toward higher throughput phenotyping. Although our study was limited to a single species and some detection challenges remain, the methodology is broadly applicable and offers practical benefits for crop phenotyping research. Results demonstrate that reducing annotation bottlenecks and addressing small object class imbalance can meaningfully enhance model effectiveness. Future work will focus on extending the framework to additional species, exploring the potential of large vision-language models and domain-specific fine-tuning strategies for specialized segmentation tasks, and integrating segmentation output into traits extraction workflow for agriculture applications.

\section*{Data availability}
\label{sec:data_availability}

The Sorghum Stomata Dataset (StomataSeg) is available for download at \url{https://zenodo.org/records/18216859} under the CC BY-NC-ND 4.0 license. The primary distribution is patch-based: the original patches are stored inside the \texttt{original\_patches/} directory, and the pseudo labels patches are stored inside the \texttt{pseudo\_labels\_patches/} directory. For completeness and reproducibility, the original full-resolution microscope images and their original COCO annotation files are provided as supplementary material in \texttt{original\_images/}. Each split (train/val/test) includes matching COCO-style JSON annotations with three instance classes (pore area, guard cell area, complex area).


\section*{Code availability}
\label{sec:code_availability}

All code supporting the methodological pipeline presented in this study is publicly available at \url{https://github.com/Davidhzt/StomataSeg_full}. The repository includes the three core modules of our approach: (1) instance-segmentation baselines built on the \href{https://github.com/open-mmlab/mmdetection}{MMDetection} framework, (2) semantic-segmentation baselines using \href{https://github.com/open-mmlab/mmsegmentation}{MMSegmentation}, and (3) the semi-supervised pseudo-labelling pipeline employed for dataset expansion. Each module is accompanied by configuration files, training scripts, and evaluation tools that replicate our experiments and facilitate further extension or adaptation.

\section*{Acknowledgements}
\label{sec:acknowledgements}

We thank the collaborators at the The University of Queensland School of Agriculture and Food Sustainability for providing valuable insights into plant physiology and stomatal biology, which strengthened the scientific basis of this work. We also acknowledge the microscopy support provided during image acquisition, including Prof Krishna Jagadish (Texas Tech University, Lubbock, Tx, USA) for providing additional sorghum images. We are grateful to the annotation team, including Xiaolong Chen, Thuong Nguyen and Ana Carolina, for their dedicated efforts in generating high-quality pixel-level labels that underpin this dataset. We also acknowledge the facilities and technical support provided by the Australian Plant Phenomics Network (APPN-UQ), supported by the Australian Government’s National Collaborative Research Infrastructure Strategy (NCRIS). Chaitanya Purushothama acknowledges funding through the International Research Training Group (IRTG), a collaboration between UQ (QAAFI and SAFS) and Justus Liebig University in Giessen, Germany, in part funded by the German Research Foundation.

\section*{CRediT authorship contribution statement}
\label{sec:author_contributions}
\textbf{Zhongtian Huang:} Methodology, Software, Validation, Writing - Original Draft, Visualization, Data Curation, Conceptualization.
\textbf{Zhi Chen:} Methodology, Validation, Writing - Review \& Editing, Resources, Supervision, Project administration, Conceptualization.
\textbf{Zi Huang:} Supervision, Writing - Review \& Editing, Project administration, Funding acquisition.
\textbf{Xin Yu:} Supervision, Writing - Review \& Editing.
\textbf{Daniel Smith:} Supervision, Writing - Review \& Editing, Conceptualization.
\textbf{Chaitanya Purushothama:} Data Curation, Investigation, Writing - Review \& Editing, Conceptualization.
\textbf{Erik Van Oosterom:} Supervision, Writing - Review \& Editing.
\textbf{Alex Wu:} Supervision, Writing - Review \& Editing.
\textbf{William Salter:} Resources, Writing - Review \& Editing.
\textbf{Yan Li:} Resources, Supervision.
\textbf{Scott Chapman:} Supervision, Writing - Review \& Editing, Project administration, Funding acquisition.

\section*{Declaration of competing interest}
\label{sec:competing_interests}

The authors declare no competing interests.

\bibliographystyle{elsarticle-harv} 
\bibliography{sample}

\end{document}